\documentclass[preprint,1p,times]{elsarticle}

\usepackage{tikz}
\usetikzlibrary{arrows.meta, positioning}

\usepackage[utf8]{inputenc}
\usepackage{ragged2e}
\usepackage{amssymb}
\usepackage{amsmath}
\usepackage{multirow}
\usepackage{graphicx}
\usepackage{booktabs} 
\usepackage{amsmath,amsfonts}
\usepackage{algorithmic}
\usepackage{algorithm}
\usepackage{array}
\usepackage[caption=false,font=normalsize,labelfont=sf,textfont=sf]{subfig}
\usepackage[table,xcdraw]{xcolor}
\usepackage{textcomp}
\usepackage[table]{xcolor}
\usepackage{longtable}
\usepackage{array}
\usepackage{mathtools}
\usepackage{titlesec}
\titlespacing*{\section}{0pt}{*0}{*0}
\usepackage{stfloats}
\usepackage{url}
\usepackage{float}
\usepackage{placeins}
\usepackage{verbatim}
\usepackage{graphicx}
\usepackage{subfig}
\usepackage{tabto}
\usepackage{hhline}
\usepackage{hyperref}
\usepackage{caption}
\usepackage[superscript]{cite} 
\usepackage{xurl}

\DeclareUnicodeCharacter{2212}{\textminus}

\usepackage{color,soul}
\usepackage{eqnarray,amsmath}
\sethlcolor{white}
\usepackage[cal=cm]{mathalpha}
\usepackage{lineno}
\usepackage{titlesec}

\titleformat{\section}[block]
  {\normalfont\Large\bfseries} 
  {} 
  {0em} 
  {} 

\titleformat{\subsection}[block]
  {\normalfont\bfseries} 
  {} 
  {0em} 
  {} 

\journal{}
\newcommand{\rev}[1]{\textcolor{black}{#1}}

\begin{document}
\newgeometry{a4paper, left=0.6in, right=0.6in, top=0.8in, bottom=0.8in}
\begin{frontmatter}

\title{A Multi-Plant Machine Learning Framework for Emission Prediction, Forecasting, and Control in Cement Manufacturing}

\author[1]{Sheikh Junaid Fayaz} 
\author[2]{Nestor D. Montiel-Bohorquez} 
\author[3]{Wilson Ricardo Leal da Silva} 
\author[1]{Shashank Bishnoi} 
\author[2]{Matteo Romano} 
\author[2]{Manuele Gatti} 
\author[1,4,*]{N. M. Anoop Krishnan} 

\affiliation[1]{organization={Civil and Environmental Engineering, Indian Institute of Technology Delhi},
            city={Hauz Khas},
            postcode={110016}, 
            state={New Delhi},
            country={India}}

\affiliation[2]{organization={Politecnico di Milano, Department of Energy},
            addressline={Lambruschini 4A, 20156}, 
            city={Milan},
            country={Italy}}

\affiliation[3]{organization={Fuller Technologies},
            city={2450 Copenhagen},
            country={Denmark}}

\affiliation[4]{organization={Yardi School of Artificial Intelligence, Indian Institute of Technology Delhi},
            city={Hauz Khas},
            postcode={110016}, 
            state={New Delhi},
            country={India}}
            
\affiliation[*]{Corresponding author: NMAK (krishnan@iitd.ac.in)
}

\begin{abstract}
Cement production is among the largest contributors to industrial air pollution, emitting $\sim$3 Mt NO\textsubscript{x}/year. The industry-standard mitigation approach, selective non-catalytic reduction (SNCR), exhibits low NH\textsubscript{3} utilization efficiency, resulting in operational inefficiencies and increased reagent costs. Here, we develop a data-driven framework for emission control using large-scale operational data from four cement plants worldwide. Benchmarking nine machine learning architectures, we observe that prediction error varies $\sim$3–5x across plants due to variation in data richness. Incorporating short-term process history nearly triples NO\textsubscript{x} prediction accuracy, revealing that NO\textsubscript{x} formation carries substantial process memory, a timescale dependence that is absent in CO and CO\textsubscript{2}. Further, we develop models that forecast NO\textsubscript{x} overshoots as early as nine minutes, providing a buffer for operational adjustments. The developed framework controls NO\textsubscript{x} formation at the source, reducing NH\textsubscript{3} consumption in downstream SNCR. Surrogate model projections estimate a $\sim$34–64\% reduction in NO\textsubscript{x} while preserving clinker quality, corresponding to a reduction of $\sim$290 t NO\textsubscript{x}/year and $\sim$58,000 USD/year in NH\textsubscript{3} savings. This work establishes a generalizable framework for data-driven emission control, offering a pathway toward low-emission operation without structural modifications or additional hardware, with potential applicability to other hard-to-abate industries such as steel, glass, and lime.
\end{abstract}

\end{frontmatter}
\section*{Introduction}
\noindent With increasing cement demand and global decarbonization efforts like the UN Net-Zero 2050\cite{deutch2020net}, the cement industry stands at a critical environmental crossroads. Cement production has nearly doubled over the past two decades, from $\sim$2~billion~t/yr in the early 2000s to $\sim$4~billion~t/yr in the mid-2020s \cite{ECRA2017, Andrew2019}. However, this rapid expansion has resulted in a proportional increase in the environmental burden. Annual CO$_2$ emissions from cement manufacturing have almost doubled from $\sim$0.7~billion~t/yr to $\sim$1.6~billion~t/yr over the same period; the sector now accounts for roughly 8\% of total anthropogenic CO$_2$ emissions \cite{Andrew2019, IPCC2022}. Moreover, the global cement demand is projected to reach $\sim$5.5~billion~t/yr by 2030, implying a further rise in CO$_2$ emissions in the absence of effective mitigation strategies \cite{IEA2023}. Beyond CO$_2$, cement production is also among the largest stationary industrial sources of nitrogen oxides (NO\textsubscript{x}), which features among key environmental performance indicators of cement kilns \cite{EUBREF2013}. NO\textsubscript{x} emissions represent the dominant short- to medium-term pollutant from cement production, driving tropospheric ozone formation, secondary particulate matter, vegetation damage, and perturbations in atmospheric chemistry \cite{WHO2021}. Therefore, developing scalable solutions for emission control is crucial for environmentally sustainable growth of the sector \cite {xu2022control}.\\
\indent NO\textsubscript{x} formation in cement kilns arises from tightly coupled thermal and chemical processes occurring primarily in the burning zone. In this zone, key factors governing NO\textsubscript{x} kinetics---including flame temperature, local stoichiometry, radical concentrations, and gas-phase residence times near the burner---cannot be reliably measured in full-scale operation \cite{zheng2020modeling}. These regions are characterized by extreme operating conditions, with flame temperatures typically exceeding 1500-2000\textdegree{C}, rendering direct instrumentation infeasible and measurements unreliable \cite{glassman2014combustion, Turns2012}. As a consequence, physics-based NO\textsubscript{x} models require detailed (and often unavailable) process information, are computationally intensive, and remain poorly suited for real-time prediction or control in industrial kilns.\\
\indent Traditionally, NO\textsubscript{x} emissions are controlled in  cement kilns through primary measures that suppress thermal NO\textsubscript{x} formation at the source---including staged combustion, flame cooling, and air-fuel ratio control---alongside secondary post-combustion mitigation strategies, such as selective non-catalytic reduction (SNCR) and selective catalytic reduction (SCR)\cite{zheng2020modeling,okoji2023evaluation}. While primary measures are capable of delivering 10-50\% NO\textsubscript{x} reduction~\cite{guseva2021nitrogen}, their effectiveness is highly configuration-dependent. Staged combustion is most effective in wet or long dry kilns, offering limited flexibility in modern precalciner systems \cite{ECRA2017}, and carries the risk of increased CO emissions. In contrast, post-combustion strategies offer substantially higher NO\textsubscript{x} reduction with 30-90\% for SNCR and 43-95\% for SCR \cite{guseva2021nitrogen}. However, the deployment of SCR requires high capital expenditure~\cite{Li2018, Tan2019}, a large installation space, catalyst deactivation, and offers limited suitability for retrofitting. This renders them impractical for many existing plants \cite{EUBREF2013}. Also, poorly managed SCR operation may promote CO formation, thereby adversely affecting clinkerization\cite{Dynamis2025}. On the other hand, while SNCR systems are comparatively inexpensive, they impose operational challenges, including a narrower, higher operating temperature window than SCR \cite{Dong2012, Li2014}, and a low efficiency of ammonia (NH$_3$) utilization, leading to evaporation of unreacted NH$_3$, namely  ammonia slip \cite{EPA_SNCR}. Although downstream oxidation catalysts can mitigate ammonia slip, they further increase system complexity and operating costs \cite{Hansen2017}. Moreover, high SNCR reduction rates typically require aggressive NH$_3$ dosing, amplifying the risk of NH$_3$ overshoots\cite{guseva2021nitrogen}. These limitations highlight the inherent challenges with relying on downstream-only mitigation solutions and underscore the need for complementary upstream, process-integrated NO\textsubscript{x} control strategies.\\
\indent Due to the limitations of traditional control and physics-based modelling, several data-driven methods for NO\textsubscript{x} analysis in cement kilns have been proposed. However, the majority of these studies focus on NO\textsubscript{x} prediction rather than NO\textsubscript{x} control~\cite{hao2021prediction, Li2004,Liukkonen2012,li2024nox}. The existing studies on NO\textsubscript{x} reduction rely on small-scale datasets spanning only a few hours to several weeks~\cite{zheng2020modeling,okoji2023evaluation,xu2022control}, thereby failing to capture the wide operational variability of a full-scale industrial kiln. Besides, temporal effects are commonly neglected, despite the inherently evolving thermal and chemical fields governing NO\textsubscript{x} formation~\cite{usman2024prediction,zheng2020modeling,okoji2023evaluation,xu2022control,hao2021prediction}. Studies on NO\textsubscript{x} control lack domain-informed constraints, which may yield mathematically optimal yet operationally infeasible solutions, and rarely validate optimised kiln configurations against the impact on production throughput or end-product quality arising from operational reconfiguration. In addition, control performance is evaluated only under low-NO\textsubscript{x} conditions (typically $<250$~PPM), leaving effectiveness under high-emission scenarios unexamined. Finally, studies~\cite{usman2024prediction} often rely on synthetic data generated from process simulations or computational fluid dynamics (CFD) simulations rather than real plant data, making their applicability for continuous industrial-scale operations in real-world questionable.\\
\indent  Given the challenges with existing NO\textsubscript{x} mitigation approaches and the growing availability of operational data due to industrial digitization activities, a crucial question arises: can a data-driven framework be developed to control NO\textsubscript{x} emissions in existing full-scale cement kilns while maintaining production targets? To this end, we curate large-scale operational data from four full-scale operational cement plants spanning multiple continents and diverse kiln configurations to perform a global, multi-plant benchmark of nine machine learning (ML) architectures for predicting NO\textsubscript{x}, CO, and CO$_2$ emissions. Rather than relying solely on instantaneous kiln states, we incorporate short-term historical evolution of key processes, enabling the models to capture the dynamic nature of kiln operation in emission prediction. Building on these predictive models, we develop an optimization framework for NO\textsubscript{x} control. Using ML-based surrogate predictions validated against historically observed operating conditions via closest-match validation, we estimate an average NO\textsubscript{x} reduction of $\sim$34--64\%. These projected reductions are comparable to state-of-the-art SNCR and SCR systems, without incurring their associated capital and operating costs. Crucially, since inherent time lags between control actions and observable NO\textsubscript{x} responses render abrupt interventions ineffective, we develop early-warning forecasters that flag high-emission episodes several minutes in advance, enabling stable control through gradual operational reconfiguration. Altogether, we propose a process-integrated framework that controls NO\textsubscript{x} formation at the source, thereby reducing reagent consumption in downstream mitigation and avoiding capital-intensive retrofits. \\
\section{Results} 
\subsection{Global multi-plant benchmark for modelling NO\textsubscript{x} and CO emissions}
\label{subsec:benchmarking}
\begin{figure}[H]
\resizebox{\textwidth}{!}{%
    \centering
    \includegraphics[scale=1.0]{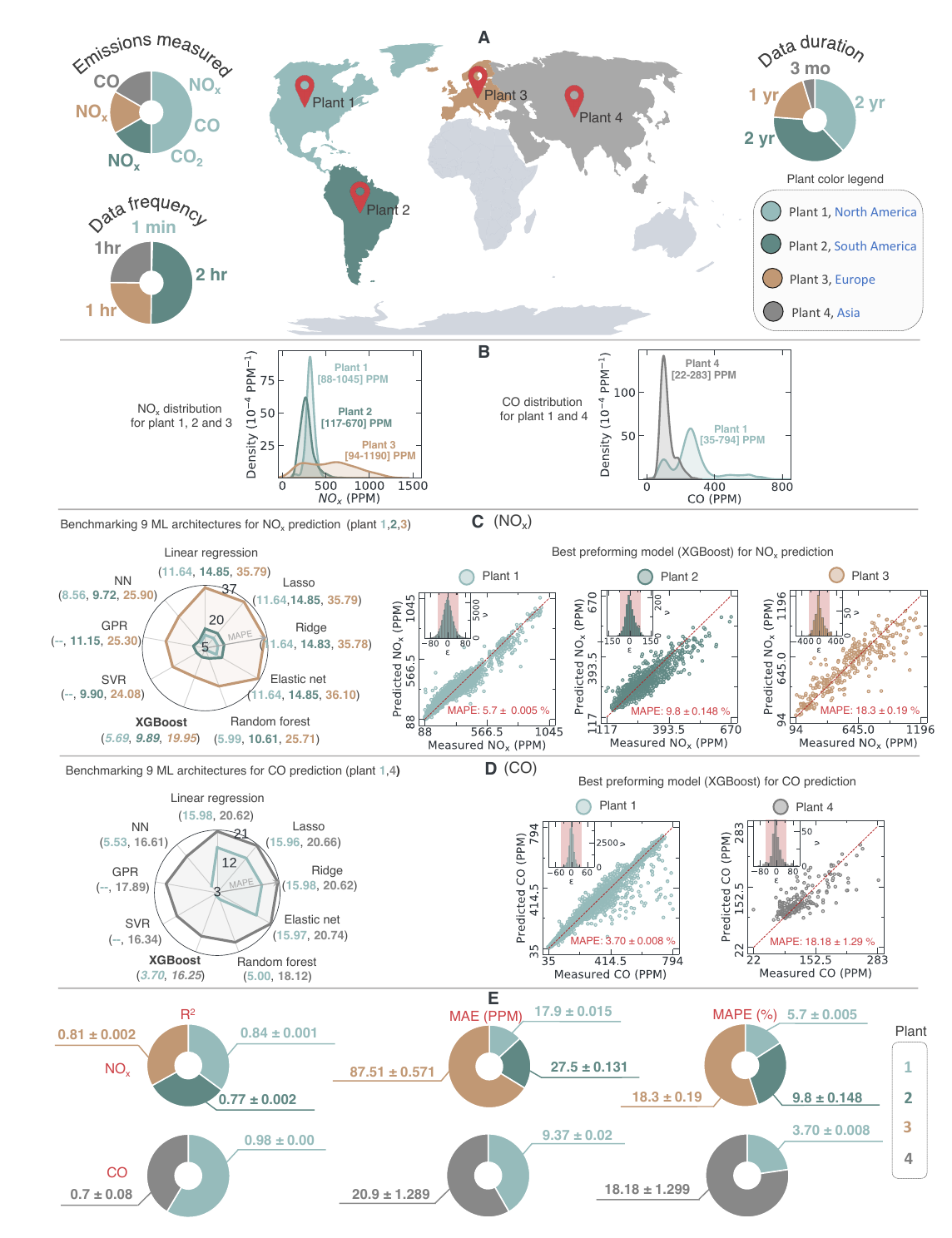}}
    \caption{\textbf{Benchmarking nine ML architectures for predicting NO\textsubscript{x} and CO across four plants.} \textbf{(A)} Plant locations with details of data collected. \textbf{(B)} Comparison of pre-processed data distributions for the four plants. Performance comparison of nine ML architectures for \textbf{(C)}  NO\textsubscript{x} prediction across plants 1-3 and \textbf{(D)} CO prediction across plants 1 and 4. Radar plots show colour-coded MAPE scores (as per the plant colour legend) on the test sets; the best-performing architecture with the lowest MAPE is highlighted in bold. Parity plots are shown for the best-performing architectures for predicting NO\textsubscript{x} (Plants 1-3) and CO (Plants 1 and 4). Inset histograms show the error distributions ($\epsilon$ = predicted - actual); red-shaded regions denote 95\% confidence intervals ($\pm$2 $\sigma$), with x-axis limits set to 99.9\% confidence ($\pm$4 $\sigma$). \textbf{(E)} Comparison of error metrics for NO\textsubscript{x} and CO prediction across plants. The circular arcs visualize the relative magnitudes only and do not necessarily sum to 100\%.}
    \label{plant_compare_fig}
\end{figure}

\noindent Industrial emission prediction (EP) models are often developed and evaluated using data from a single facility, implicitly assuming similar predictive capabilities across different kiln configurations and fuel mixes. Despite broader agreement that this may not be true, the same has not been rigorously evaluated, often due to paucity of diverse industrial data. Here, we curate large-scale operational datasets from four full-scale cement plants located in North America, South America, Europe, and Asia, exhibiting diverse kiln architectures, fuel mixes, data volumes, and sampling frequencies as described in Fig.~\ref{plant_compare_fig}A and Table ~\ref{plant_compare_table}. CO$_2$ measurement is available only for Plant~1 and is therefore excluded from the multi-plant benchmarking. Accordingly, NO\textsubscript{x} models are benchmarked using data from Plants~1--3, while CO models are benchmarked using data from Plants~1 and~4. Figure~\ref{plant_compare_fig}B compares the distributions of the pre-processed emissions (data curation and pre-processing protocol described in Methods). For NO\textsubscript{x}, Plant~3 exhibits a more dispersed distribution, whereas Plants~1 and~2 show higher data density around $\sim$250~ppm with all plants exhibiting comparable concentration ranges. For CO, Plant~1 displays a wide distribution across its operating range, while Plant~4 is more narrowly concentrated around $\sim$100~ppm. Leveraging this diversity in plant configurations and emission distributions, we benchmark nine machine-learning architectures, namely, linear regression, lasso, ridge, elastic net, random forest, XGBoost, support vector regression (SVR), Gaussian process regression (GPR), and neural networks (NN), for predicting NO\textsubscript{x} and CO emissions. Mathematical construction of all these architectures is discussed in Supplementary~C.

\begin{table}[H]
\centering
\caption{\textbf{Plant features:} Comparison of configurations, fuel mixes, emission monitoring, and details of raw and pre-processed data across the four plants.}
\label{plant_compare_table}
\resizebox{\columnwidth}{!}{%
\begin{tabular}{|cccc|cccccccccc|}
\hline
\multicolumn{4}{|c|}{\textbf{Cement plant specifications}} &
  \multicolumn{10}{c|}{\textbf{Data specifications}} \\ \hline
\multirow{2}{*}{\textbf{\begin{tabular}[c]{@{}c@{}}\\\\  ID\end{tabular}}} &
  \multirow{2}{*}{\textbf{\begin{tabular}[c]{@{}c@{}}\\\\ Location\end{tabular}}} &
  \multirow{2}{*}{\textbf{\begin{tabular}[c]{@{}c@{}}\\\\Configuration\end{tabular}}} &
  \multirow{2}{*}{\textbf{\begin{tabular}[c]{@{}c@{}}\\\\ Fuel mix\end{tabular}}} &
  \multirow{2}{*}{\textbf{\begin{tabular}[c]{@{}c@{}}Emissions \\ measured\\(PPM) \end{tabular}}} &
  \multirow{2}{*}{\textbf{\begin{tabular}[c]{@{}c@{}}\\Measurement\\  location\end{tabular}}} &
  \multirow{2}{*}{\textbf{\begin{tabular}[c]{@{}c@{}}\\Sampling\\  frequency\end{tabular}}} &
  \multicolumn{1}{c|}{\multirow{2}{*}{\textbf{\begin{tabular}[c]{@{}c@{}}\\Data\\ Duration\end{tabular}}}} &
  \multicolumn{4}{c|}{\textbf{Size \textcolor{blue}{$\ddagger$}}} &
  \multicolumn{2}{c|}{\textbf{\begin{tabular}[c]{@{}c@{}}Process \\ parameters\end{tabular}}} \\ \cline{9-14}
 &
   &
   &
   &
   &
   &
   &
  \multicolumn{1}{c|}{} &
  \textbf{Raw} &
  \textbf{\begin{tabular}[c]{@{}c@{}}After\\  consistency \\ check\end{tabular}} &
  \textbf{\begin{tabular}[c]{@{}c@{}}After \\ physical \\ validation\end{tabular}} &
  \multicolumn{1}{c|}{\textbf{\begin{tabular}[c]{@{}c@{}}After \\ outlier \\ removal\end{tabular}}} &
  \textbf{\begin{tabular}[c]{@{}c@{}}Given\\  in raw \\ data\end{tabular}} &
  \textbf{\begin{tabular}[c]{@{}c@{}}Used\\  in \\ study\end{tabular}} \\ \hline
\textbf{1} &
  North America &
  \begin{tabular}[c]{@{}c@{}}5 stage PHT \\ +\\ ILC pre-calciner\end{tabular} &
  Coal &
  \begin{tabular}[c]{@{}c@{}}NO$_{x}$\\ CO\\ CO$_{2}$\end{tabular} &
  \begin{tabular}[c]{@{}c@{}}kiln inlet\textcolor{blue}{$\dagger$} (NO$_{x}$)\\ stack end (CO)\\  stack end (CO$_{2}$)\end{tabular} &
  1 minute &
  \multicolumn{1}{c|}{\begin{tabular}[c]{@{}c@{}}01 Jan 20 --\\ 31 Dec 21\\   (2 year)\end{tabular}} &
  1,052,567 &
  894,036 &
  277,603 &
  \multicolumn{1}{c|}{277,079} &
  44 &
  40 \\ \hline
\textbf{2} &
  South America &
  \begin{tabular}[c]{@{}c@{}}5 stage PHT \\ +\\ ILC pre-calciner\end{tabular} &
  \begin{tabular}[c]{@{}c@{}}Petcoke,\\ charcoal,\\ waste wood\\ chip,\\ corn husk,\\ rice husk, \\ residual tires\end{tabular} &
  NO$_{x}$ &
  kiln inlet &
  2 hour &
  \multicolumn{1}{c|}{\begin{tabular}[c]{@{}c@{}}11 Jan 20 --\\ 31 Dec 21\\ (2 year)\end{tabular}} &
  6,907 &
  6,907 &
  6,907 &
  \multicolumn{1}{c|}{5,994} &
  47 &
  18 \\ \hline
\textbf{3} &
  Europe &
  \begin{tabular}[c]{@{}c@{}}Reinforced \\ suspension \\ preheater (RSP) \\ tower\end{tabular} &
  \begin{tabular}[c]{@{}c@{}}Petcoke, \\ RDF\end{tabular} &
  NO$_{x}$ &
  kiln inlet &
  1 hour &
  \multicolumn{1}{c|}{\begin{tabular}[c]{@{}c@{}}30 Nov 21 --\\ 22 Dec 22\\ (1 year)\end{tabular}} &
  5,784 &
  2,621 &
  2,621 &
  \multicolumn{1}{c|}{1,620} &
  74 &
  31 \\ \hline
\textbf{4} &
  Asia &
  \begin{tabular}[c]{@{}c@{}}6 stage PHT \\ +\\ ILC pre-calciner\end{tabular} &
  \begin{tabular}[c]{@{}c@{}}Petcoke, \\ coal\end{tabular} &
  CO &
  stack end &
  1 hour &
  \multicolumn{1}{c|}{\begin{tabular}[c]{@{}c@{}}01 Feb 22 --\\ 01 May 22\\  (3 months)\end{tabular}} &
  2,165 &
  1,389 &
  1,217 &
  \multicolumn{1}{c|}{1,151} &
  127 &
  54 \\ \hline
\end{tabular}%
}
\vspace{0em}
\footnotesize
\raggedright
\textsuperscript{\textcolor{blue}{$\dagger$}}\ \textit{Kiln inlet refers to the calciner end of the rotary kiln} \\
\textsuperscript{\textcolor{blue}{$\ddagger$}}\ \textit{Dataset size refers to number of samples of each parameter} \\
PHT: Preheater tower, ILC: in-line calciner
\end{table}

\noindent Figures~\ref{plant_compare_fig}C show the mean absolute percentage error (MAPE) for NO\textsubscript{x} predictions across Plants~1--3. Linear models consistently exhibit the highest errors, reflecting their limited capacity to capture the non-linear combustion dynamics inherent to cement kilns (see Figs.~S16--S17). Across all three plants, XGBoost consistently achieves the lowest MAPE for NO\textsubscript{x} prediction. A similar trend is observed for CO prediction (Fig.~\ref{plant_compare_fig}D), with linear models underperforming non-linear approaches, and XGBoost yielding the lowest MAPE for both Plants~1 and~4. Figures~\ref{plant_compare_fig}C and~D illustrate the performance of the best model (XGBoost) for NO\textsubscript{x} and CO prediction across plants. Notably, SVR and GPR could not be evaluated for Plant~1 due to the prohibitive memory requirements arising from the need to allocate a covariance matrix of approximately 128~GB for datasets exceeding $10^6$ samples. In Fig. \ref{plant_compare_fig}E, the prediction errors for the best performing models (XGBoost) follow a distinct order:
\begin{align*}
\text{MAPE (NO\textsubscript{x}): } \quad & \mathrm{5.7\pm0.005\% }\text{ (Plant 1)}
< \mathrm{9.8\pm0.148\% }\text{ (Plant 2)}
< \mathrm{18.3\pm0.019\% }\text{ (Plant 3)},
\end{align*}
\begin{align*}
\text{MAPE (CO):} \quad & \mathrm{3.7\pm0.008\%}\text{ (Plant 1)} \ll \mathrm{18.2\pm1.29\%}\text{ (Plant 4)}
\end{align*}
Despite employing identical pre-processing, hyperparameter tuning, and training protocols across all plants (Supplementary~D1--D4), model accuracy varies systematically with data volume and temporal granularity. High-volume and high-frequency datasets (e.g., Plant~1 with 2 years of data at 1-min resolution) enable XGBoost to learn accurate, stable process-emission relationships. In contrast, low-volume and coarse-resolution datasets (e.g., Plant~4 with three months of data at 1-h resolution) fundamentally limit achievable accuracy. The above trend underscores the importance of data quality and richness for developing accurate industrial-scale EP models.

\subsection{EP vs. emission prediction with historical inputs (EPH) for modelling NO\textsubscript{x}, CO, and CO$_2$ }
\label{subsec:steady_vs_process_evolution}
\begin{figure}[H]
\resizebox{\textwidth}{!}{%
    \centering
    \includegraphics[scale=1.0]{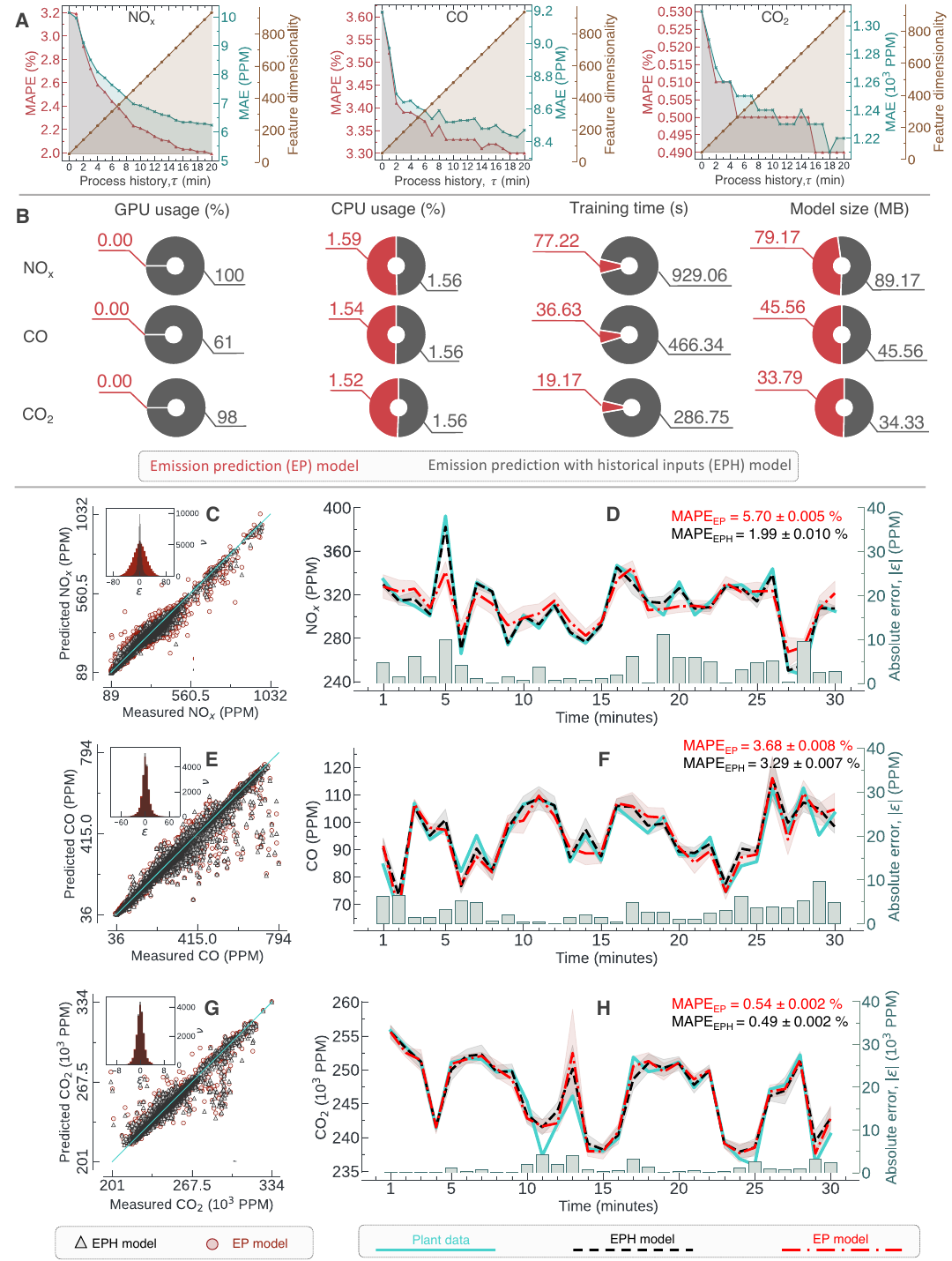}}
\caption{\textbf{EP vs EPH models for Plant 1.} (\textbf{A}) Error vs. process history for the EPH model. (\textbf{B}) Comparison of computational load; circular arcs denote relative magnitudes and do not necessarily sum to 100\%. (\textbf{C--H}) Performance of EP and EPH models for predicting NO$_{x}$, CO, and CO$_{2}$. Inset histograms in the parity plots show error distributions ($\epsilon$ = predicted$-$actual) on the test set with x-axis limits set to the 99.9\% confidence range ($\pm4\sigma$). The inset and parity plot share the same units. Right subplots (\textbf{D, F, H}) show 30-min temporal predictions against plant data: solid lines denote mean predictions, shaded bands represent model uncertainty ($\pm3\sigma$), and green bars (right axis) indicate absolute prediction errors. Results over the full two-year dataset are shown in Fig.~S24 (Supplementary~E).}

    \label{current_vs_process_fig}
\end{figure}

\noindent Since kiln operations span multiple time-scales, plant emissions depend not only on instantaneous operating states but also on the temporal evolution of process conditions. To quantify the influence of prior operation on emission dynamics, we compare EP and EPH modelling approaches for predicting NO\textsubscript{x}, CO, and CO$_2$. The EP models evaluated in the preceding benchmarking section assume that the emission at a given time is a function of the process parameters (PP) at that time, given by:
\begin{equation}
\hat{y}(t) = f\big(\mathbf{PP}(t)\big).
\label{eq:steady_state}
\end{equation}
This equation neglects transient effects arising from historical variations in combustion, mixing, or plant operation. In contrast, EPH model incorporates short-term process memory by conditioning predictions on a finite history of process parameters (PP) as:
\begin{equation}
\hat{y}(t) = f\big(\mathbf{PP}(t), \mathbf{PP}(t-1), \ldots, \mathbf{PP}(t-\tau)\big),
\label{eq:process_evolution}
\end{equation}
where $\tau$ denotes the process history-dependent length. The value of $\tau$ governs a fundamental trade-off between predictive accuracy and model complexity. In this study, we select $\tau$ to be 20~min (Fig. \ref{current_vs_process_fig}A). The systematic procedure for selecting $\tau$ is described in Process-history selection (see Methods).

\noindent Figure~\ref{current_vs_process_fig}B shows the computational overhead for developing EPH models. Despite the substantial increase in input dimensionality, EP and EPH models exhibit comparable CPU utilization ($\sim$1.52\%) and identical model sizes, as both share the same architecture and hyperparameters (Tables S8--S10) for a given emission species. The computational distinction emerges during training: EPH models require GPU acceleration and incur an order-of-magnitude increase in training time, from 77~s to 930~s for NO\textsubscript{x}, 36~s to 467~s for CO, and 19~s to 286~s for CO$_2$. Inference times, however, remain identical. Train vs test performance of both models is provided in Table~S11 and Fig. S23, Supplementary~E.
\noindent Figure~\ref{current_vs_process_fig}B shows the computational overhead for developing EPH models. Despite the substantial increase in input dimensionality, EP and EPH models exhibit comparable CPU utilization ($\sim$1.52\%) and identical model sizes, as both share the same architecture and hyperparameters (Tables S8--S10) for a given emission species. The computational distinction emerges during training: EPH models require GPU acceleration and incur an order-of-magnitude increase in training time, from 77~s to 930~s for NO\textsubscript{x}, 36~s to 467~s for CO, and 19~s to 286~s for CO$_2$. Inference times, however, remain identical.

Comparing the EP and EPH-based prediction for NO\textsubscript{x} (Figs.~\ref{current_vs_process_fig}C-D), CO (Figs.~\ref{current_vs_process_fig}E-F), and CO$_2$ (Figs.~\ref{current_vs_process_fig}G-H), a clear emission-specific response to including temporal history is revealed. CO and CO$_2$ predictions show only marginal improvement when process history is included:
\begin{align*}
\text{CO:} \quad & \text{MAPE} = 3.68\pm0.008\% \;\text{(EP)} \;\; \text{vs.} \;\; 3.29\pm0.007\% \;\text{(EPH)}, \\
\text{CO\textsubscript{2}}: \quad & \text{MAPE} = 0.54\pm0.002\% \;\text{(EP)} \;\; \text{vs.} \;\; 0.49\pm0.002\% \;\text{(EPH)}.
\end{align*}
In contrast, NO\textsubscript{x} prediction exhibits a substantial ($\sim$3 times) accuracy gain from temporal context,
\begin{align*}
\text{NO\textsubscript{x}: } \quad & \text{MAPE} = \; 5.70\pm 0.005\% \;\text{(EP)} \;\; \text{vs.} \;\; 1.99\pm0.010\% \;\text{(EPH)},
\end{align*}
This is also reflected by the error histogram in Fig.~\ref {current_vs_process_fig}C showing a narrow, near-zero-centred distribution for the EPH model, in stark contrast to the broader error spread of the EP formulation. Although NO\textsubscript{x}, CO, and CO$_2$ share the same gas-phase residence time---governed by kiln geometry, gas velocity, and draft---their effective \emph{chemical} residence times differ markedly \cite{Turns2012, Glassman2014}. NO\textsubscript{x} formation and re-equilibration persist over minutes under evolving thermal and stoichiometric conditions, imparting memory of prior operation \cite{Miller1999, Zeldovich1946}. By contrast, CO is a short-lived intermediate rapidly oxidized once favorable conditions arise, while CO$_2$ is a stable end-product reflecting cumulative fuel conversion \cite{Turns2012 , Glassman2014}. Consequently, temporal modelling substantially improves NO\textsubscript{x} prediction but yields only marginal gains for CO and CO$_2$.

It is important to mention that EPH models are developed exclusively for Plant~1, as it provides data at 1-minute resolution, enabling the construction of short-term process histories spanning several minutes. In contrast, the remaining three plants report data at coarse temporal resolutions (1-2 h), which precludes meaningful representation of minute-scale EPH. For the same temporal-granularity constraint, all forecasting and optimization analyses in subsequent sections are confined to Plant~1.

\subsection{Emission forecasting: early-warning alarm for high-emission episodes}
\label{subsec:forecast_performance}
\begin{figure}[H]
\resizebox{\textwidth}{!}{%
    \centering
    \includegraphics[scale=1.0]{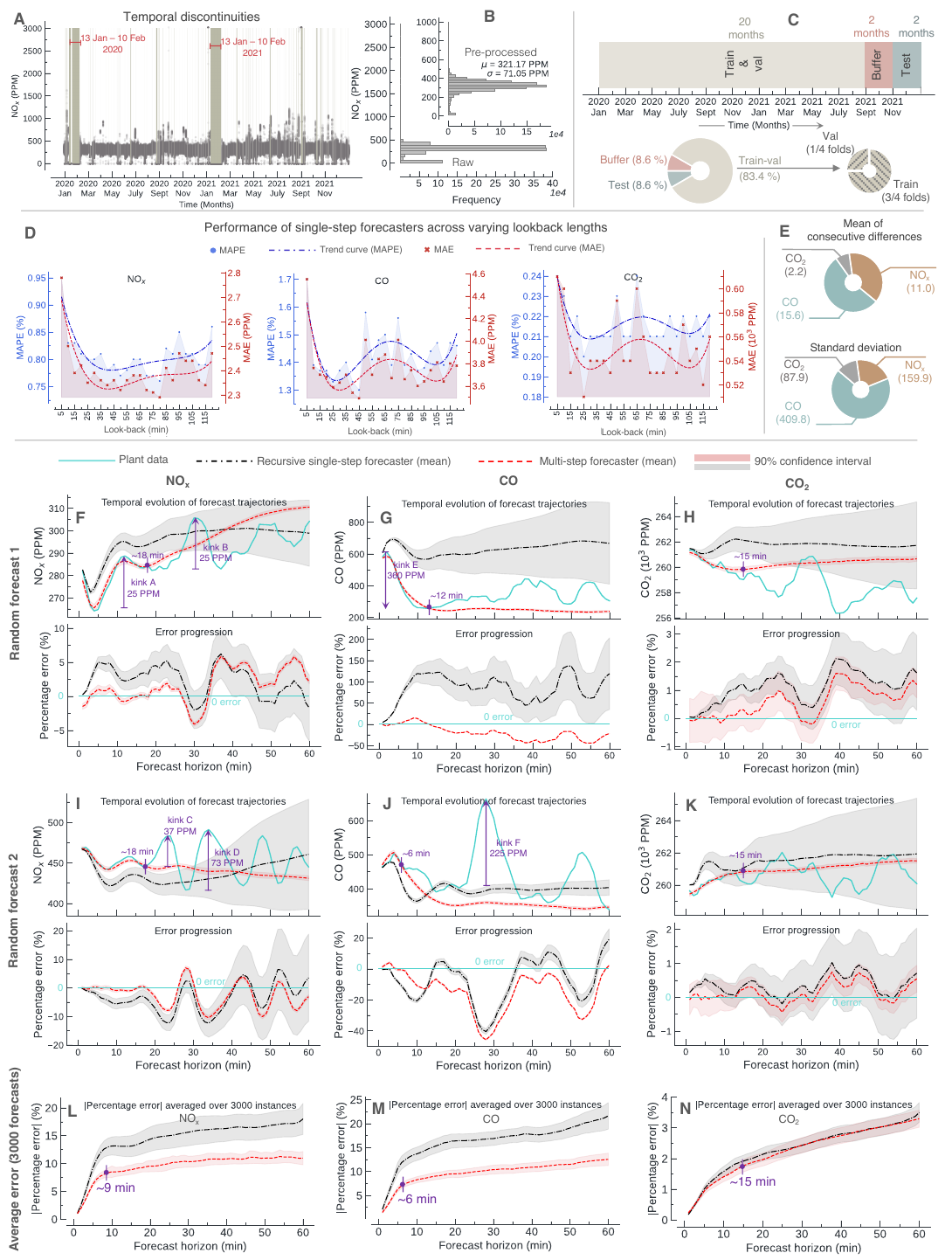}}
\caption{\textbf{Auto-regressive forecast of NO\textsubscript{x}, CO and CO$_2$.} 
(\textbf{A}) Vertical patches indicate temporal gaps in the time-series due from to missing records or pre-processing. Two major gaps occur during scheduled plant maintenance \textit{13 Jan 2020 - 10 Feb 2020} and \textit{13 Jan 2021 - 10 Feb 2021}.
(\textbf{B}) Raw and pre-processed NO\textsubscript{x} data.
(\textbf{C}) Data-split strategy for developing the forecasters. \textit{01 Jan 2020 - 30 Aug 2021} for training and 4-fold cross-validation, \textit{01 Sept 2021 - 30 Oct 2021} as temporal buffer between train and test phases, and \textit{01 Nov 2021 - 30 Dec 2021} for testing.
(\textbf{D}) MAPE and MAE vs. look-back lengths for the single-step forecasters. Forecasters are evaluated for look-backs ranging from 5 mins to 2 hrs, with a five-min increment at each iteration. MAPE (left y-axis) is shown with blue circles, and MAE (right y-axis) with red crosses; trend lines are $4^{th}$ order polynomial fits.
(\textbf{E}) Mean of consecutive differences and and the standard deviation of the processed emissions time-series.
(\textbf{F-K}) Recursive single-step vs. multi-step forecasters over a 60-minute forecast horizon for two randomly selected events: (F-H) event 1 and (I-K) event 2. Upper panels show forecast trajectories vs. plant data; lower panels show error trajectories w.r.t the zero-error . Dotted lines denote mean trajectories and shaded regions the $90\%$ confidence interval. 
(\textbf{L-N}) Absolute percentage errors averaged over 3000 random instances comparing recursive single-step and multi-step forecasters.}    
    \label{forecast_trajectories_fig}
\end{figure}

\noindent Using two years of operational emission data from Plant~1, we develop forecasters for NO\textsubscript{x}, CO, and CO$_2$. The forecasters use a 25-minute look-back window, selected as described in the Methods. They are evaluated over a 60-minute forecast horizon to examine how far into the future emission trajectories can be reliably anticipated. We compare two forecasting strategies with distinct temporal constructions (see Methods for mathematical formulations): (i) \emph{recursive single-step} forecaster, which predicts emissions one minute ahead and propagates predictions forward in time, and (ii) \emph{multi-step} forecaster, which predicts the entire 60-minute emissions in a single forward pass. To avoid data leakage, both models are developed using a chronological data split (Fig. ~\ref{forecast_trajectories_fig}C): data from 01 Jan 2020--30 Aug 2021 are used for training and cross-validation, followed by a two-month temporal buffer (01 Sept--30 Oct 2021), and two months (01 Nov--30 Dec 2021) reserved exclusively for testing the forecasters.

Figures~\ref {forecast_trajectories_fig}F-K compares recursive single-step and direct multi-step forecasts for two random forecasting events (additional events in Fig.~S32, Supplementary F5). Across all emissions, the recursive single-step forecasts (shown in black) deteriorate notably faster, diverging from the plant trajectory (green) within the first few minutes. In some cases, this approach results in forecasting physically inconsistent trends even from the very first forecast step (see Figs.~\ref{forecast_trajectories_fig}E, I, J). This behaviour arises from compounding errors (for details, see Error propagation in Methods) with recursive forecasts, which also lead to a rapid explosion of forecast uncertainty (the $90\%$ confidence interval for the forecaster), rendering them unsuitable for reliable forecasts. Therefore, the discussion henceforth focuses only on the multi-step forecasters (shown in red).

For NO\textsubscript{x}, the multi-step forecaster accurately predicts a $\sim$25~PPM fluctuation occurring within the first 18 minutes of the forecast (kink~A, Fig.~\ref{forecast_trajectories_fig}F), while failing to anticipate a fluctuation (kink~B) of the same magnitude occurring later ($\sim$30 minutes) in the same event. Similarly, in the other event (Fig.~\ref{forecast_trajectories_fig}I), fluctuations occurring beyond $\sim$18 minutes (kinks~C and~D) are also missed, suggesting an 18-minute horizon for reliable trajectory forecast. However, in some cases, the reliability of multi-step forecasting drops notably at shorter horizons, such as $\sim$12 minutes (Fig. S32B) or even $\sim$6 minutes (Fig. S32B), highlighting the inconsistency in the ability to forecast across different events. An analogous behaviour is observed for CO, albeit over different time scales. The multi-step forecaster accurately captures large-amplitude fluctuations (e.g., $\sim$360~PPM, kink~E) occurring within the first $\sim$12 minutes (Fig. ~\ref{forecast_trajectories_fig}G). However, in some cases, trajectory prediction degrades as early as $\sim$6 minutes (Fig. S~\ref{forecast_trajectories_fig}J) missing even smaller excursions (e.g., $\sim$225~PPM, kink~F) occurring later in the forecast. CO$_2$, in contrast, exhibits smoother dynamics with a more consistent 15-minute forecast limit (Figs. \ref{forecast_trajectories_fig}H, K; Figs. S32I-L).

To eliminate event-selection bias in forecasting, we calculate absolute percentage errors averaged over 3,000 randomly sampled forecast events, yielding an unbiased estimate of forecast-ability (Figs.~\ref{forecast_trajectories_fig}L-N). For NO\textsubscript{x}, the forecast error plateaus at approximately $\sim$9 minutes (Fig.~\ref{forecast_trajectories_fig}L), stabilizing near $\sim$7\%, indicating the practical limit of reliable trajectory prediction. CO exhibits earlier saturation, with the error reaching $\sim$7\% at approximately $\sim$6 minutes (Fig.~\ref{forecast_trajectories_fig}M). In contrast, CO\textsubscript{2} forecasts maintain errors below 2\% for up to $\sim$15 minutes (Fig.~\ref{forecast_trajectories_fig}N). Accordingly, the effective forecast horizons for these emissions are:
\begin{equation*}\label{forecast limit eq}
\mathrm{6}\text{ min (CO)} < \mathrm{9}\text{ min (NO\textsubscript{x})} < \mathrm{15}\text{ min (CO$_2$)}
\end{equation*}
Beyond these horizons, multi-step forecasts either collapse toward mean behaviour or diverge from the plant trajectory. Also, the difference in the historical timescales for different emissions aligns with variance in their time-series distribution (see Fig.~\ref{forecast_trajectories_fig}E) quantified using the average of absolute consecutive differences ($Avg(|\Delta y|)$ which follows the order:
\begin{equation*}\label{variance eq}
\begin{aligned}
\operatorname{Avg}\!\left(|\Delta y_{\mathrm{CO}}|\right)
&>\;
\operatorname{Avg}\!\left(|\Delta y_{NO\textsubscript{x}}|\right)
\;\gg\;
\operatorname{Avg}\!\left(|\Delta y_{\mathrm{CO}_2}|\right)\\
15.6
&>\;
11
\;\gg\;
2.2
\end{aligned}
\end{equation*}
CO exhibiting the highest variability (15.6) has the shortest forecast horizon ($\sim6$ min), and CO$_2$, having notably lower variance (2.2), allows longer forecasts up to $\sim15$ min.

\subsection*{\rev{NO\textsubscript{x} controller framework}}\label{optimize_sectiion}
\begin{figure}[H]
\resizebox{\textwidth}{!}{%
    \centering
    \includegraphics[scale=1.0]{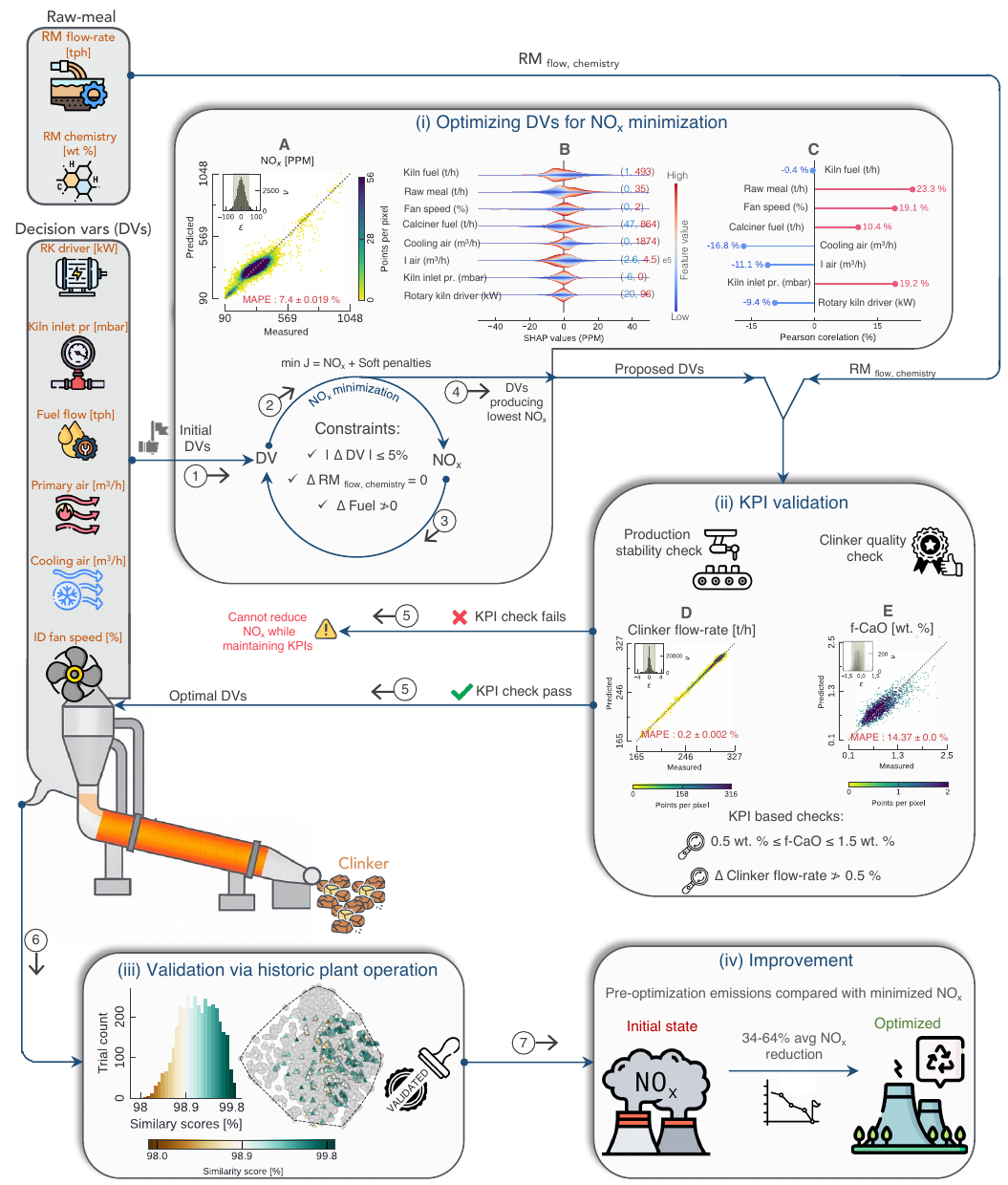}}
\caption{
\textbf{Integration of the NO\textsubscript{x} controller with plant operation.} The controller optimizes seven decision variables (DVs) to minimize NO\textsubscript{x} emissions using a surrogate ML model \textbf{(A)}. The initial DV states are derived from historical plant operation data. Post-hoc model explainability \textbf{(B-C)} ensures that ML-predicted trends are consistent with known process-emission relationships. After performing 35 optimization iterations, the controller chooses the DV combination that yields the lowest predicted NO\textsubscript{x} while satisfying the constraints: each DV can vary no more than 5\% from its initial-state value; RM flow-rate and chemistry remain fixed; and total fuel consumption must not increase. The objective function incorporates two realism-based penalty terms to discourage practically infeasible solutions.
The optimized DVs are combined with the unchanged RM parameters to predict key performance indicators (KPIs) — \textbf{(D)} clinker flow-rate and \textbf{(E)} free lime (f-CaO) — using separate ML-based surrogate models. KPI validation ensures that production stability (clinker flow-rate variation < 0.5\%) and clinker quality (f-CaO within 0.5-1.5 wt. \%) are maintained. Finally, KPI-validated optimal DVs and the corresponding minimized NO\textsubscript{x} values are compared with the closest historical operational points using Manhattan distance-based similarity scores. Across all trials, the optimized DV sets show >98 \% similarity with real plant operations, confirming their practicality, and project an average NO\textsubscript{x} reduction of 34\% over 3,761 normal operation cases.} \label{optimization_scheme_fig}
\end{figure}

\noindent Now, we present a framework that integrates the ML surrogate models and a stochastic population-based evolutionary algorithm (see Supplementary G1 for details) with plant operations to control NO\textsubscript{x} concentration levels at the kiln inlet, as shown in Fig. \ref{optimization_scheme_fig}. A multivariate process space, based on seven decision variables (DVs), is derived from historical operation data from Plant 1. DVs correspond to seven major controllable process parameters (primary air [$m^3$/h], cooling air [$m^3$/h], preheater ID fan speed [\%], calciner fuel [t/h], rotary kiln fuel [t/h], rotary kiln driver [kW], pressure at kiln inlet [mbar]) that are optimized to control the NO\textsubscript{x} emissions while raw meal flow rate [t/h] and feed chemistry [wt. \%] are held constant. In the given plant, kiln inlet pressure is a controllable parameter maintained at a setpoint by the kiln control system through adjustments in the ID fan speed. However, the relationship between fan speed and pressure is not one-to-one; the same fan speed can produce different inlet pressures depending on fuel rate, false air ingress, and RM feed. Therefore, ID fan speed and kiln inlet pressure are retained independently in the DV space to represent the coupled yet adjustable operating degrees of freedom. The controller framework is implemented through the following 3 stages.\\
\textbf{(i) Optimizing DV for NO\textsubscript{x} control:}
The actual plant-data-based values of DVs are used to initialize the optimization algorithm, which iteratively implements changes in DVs to reduce NO\textsubscript{x} until convergence is reached. Upon convergence, the controller chooses the combination of DVs producing the lowest predicted NO\textsubscript{x} while satisfying the following constraints: \\
(a) $|\Delta \mathrm{DV}| \le 5\%$: Each DV is allowed to vary within $\pm$5\% limits from the initial value, ensuring that the optimization does not entail operationally unfeasible changes in the DV.\\
(b) $\Delta \mathrm{RM}_{\text{flow-rate, chemistry}} = 0$: RM flow-rate and chemistry remain fixed because RM changes require adjustments in material sourcing, handling, and proportioning systems, thereby causing operational interruptions.\\
(c) $\Delta Fuel \le 0$: Total fuel consumption must not increase.\\
Details on the above constraints are provided in Optimization constraints (see Methods section). The optimization is expressed as a constrained non-linear minimization problem:

\begin{equation}
    J(DV) = \operatorname*{min}_{DV} \left(f_{\text{NO}_\text{x}}(DV) 
    + \underbrace{W_{\text{corr}} \, P_{\text{corr}}(DV)
    + W_{\text{operate}} \, P_{\text{operate}}(DV)}_\text{Penalties promoting operational plausibility}\right) \quad\text{subject to}\quad  
    \left\{
    \begin{aligned}
         & |\Delta \mathbf{DV}| &\le& 5\% \\
         & \Delta \mathrm{RM}_{\text{flow-rate, chemistry}} &=& 0  \\
         & \Delta \mathrm{Fuel} &\le& 0
    \end{aligned}
    \right\}
\label{Loss_func_svm}
\end{equation}
where:\\
$f_{NO\textsubscript{x}}(DV)$: ML predictions for NO\textsubscript{x} emissions.\\
$P_{corr}(DV)$: Correlation consistency penalty that penalizes candidate solutions not aligning with the correlation structure of the historic plant data, therefore preserving the correlation features.\\
$P_{operate}(DV)$: Operational feasibility penalty to promote candidate solutions falling close to the operating manifold of the plant based on the historic data.\\
$W_{corr}$ and $W_{operate}$: scalar penalty weights.\\
The mathematical construction of these penalties is given in Methods.

Note that the NO\textsubscript{x} prediction during optimization iterations is generated by an ML surrogate trained on plant data, which uses the 7 DV as input, achieving a mean absolute percentage error of 7.4 $\pm$ 0.019\% (Fig. \ref{optimization_scheme_fig}A; Supplementary G3). Moreover, we perform post-hoc explainability analysis using SHAP (SHapley Additive exPlanations) to confirm if the surrogate model captures physically meaningful process-emission relationships.

Interestingly, we find that that the SHAP based directional impact (Fig. \ref{optimization_scheme_fig}B) of DVs on the ML predictions is broadly consistent with the process-emission relationships derived independently from plant data (Fig. \ref{optimization_scheme_fig}C) using Pearson's correlations. For instance, increase in calciner fuel flow rate, preheater ID fan speed, and kiln inlet speed are statistically associated with higher predicted NO\textsubscript{x} concentrations, consistent with the positive correlations of these DVs with NO\textsubscript{x} formation. This agreement (see Table S17 for details) establishes further credibility of the ML surrogate and demonstrates how data-driven methodologies can transform traditional industrial processes while preserving interpretability and fundamental physical insights. However, it is crucial to highlight that interpreting the role of DVs towards NO\textsubscript{x} formation in full-scale cement kilns is inherently non-trivial due to the strong coupling between air and fuel flows, closed-loop draft control, and non-monotonic dependence of NO\textsubscript{x} on local stoichiometry and temperature. Accordingly, SHAP-derived influences should be interpreted as statistical attributions conditioned on historically observed operating patterns rather than as direct causal relationships.

\textbf{(ii) Key performance indicator (KPI) validation:}
DV configurations that minimise predicted NO\textsubscript{x} while satisfying the optimisation constraints are subsequently subjected to KPI validation to ensure preservation of production performance. Specifically, clinker throughput stability is assessed by enforcing a clinker flow-rate variation within $\pm 0.5$, and clinker quality is evaluated by maintaining f-CaO within 0.5--1.5 wt. \%. The optimized DVs, combined with unchanged RM flow rate and chemistry, are passed to independent ML surrogate models trained to predict clinker flow rate and f-CaO (Fig. \ref{optimization_scheme_fig}D-E; Supplementary G4-G5). DV solutions that satisfy both KPI criteria are retained, while candidates that compromise throughput or quality are discarded. Across 3,959 control trials spanning 2 years of plant operation, KPI validation fails in only 21 cases (i.e., a 0.5\% failure rate). In other words, in 99.5 out of 100 trials, the control is successful. 

\textbf{(iii) Validation against historic plant operation:}
To assess whether KPI-validated optimal DVs correspond to realizable emission reductions under real operating conditions, we compare each optimized solution against historically observed plant states with similar operating conditions. Similarity is quantified using Manhattan-distance-based metrics computed over the seven DVs and the associated NO\textsubscript{x} emissions, thereby evaluating proximity in both the control space and the emission response (see Methods for the mathematical implementation). Across all control trials, optimized DVs exhibit 98\% similarity with real historical operating points, demonstrating that the controller's recommendations are not extrapolative artefacts but lie within the envelope of previously realized plant behaviour. This confirms the practical deployability of the controller recommendations.

\subsection*{\rev{Performance of NO\textsubscript{x} controller across different operational scenarios}}
\label{optimize_results_sec}
\begin{figure}[H]
\resizebox{\textwidth}{!}{%
    \centering
    \includegraphics[scale=1.0]{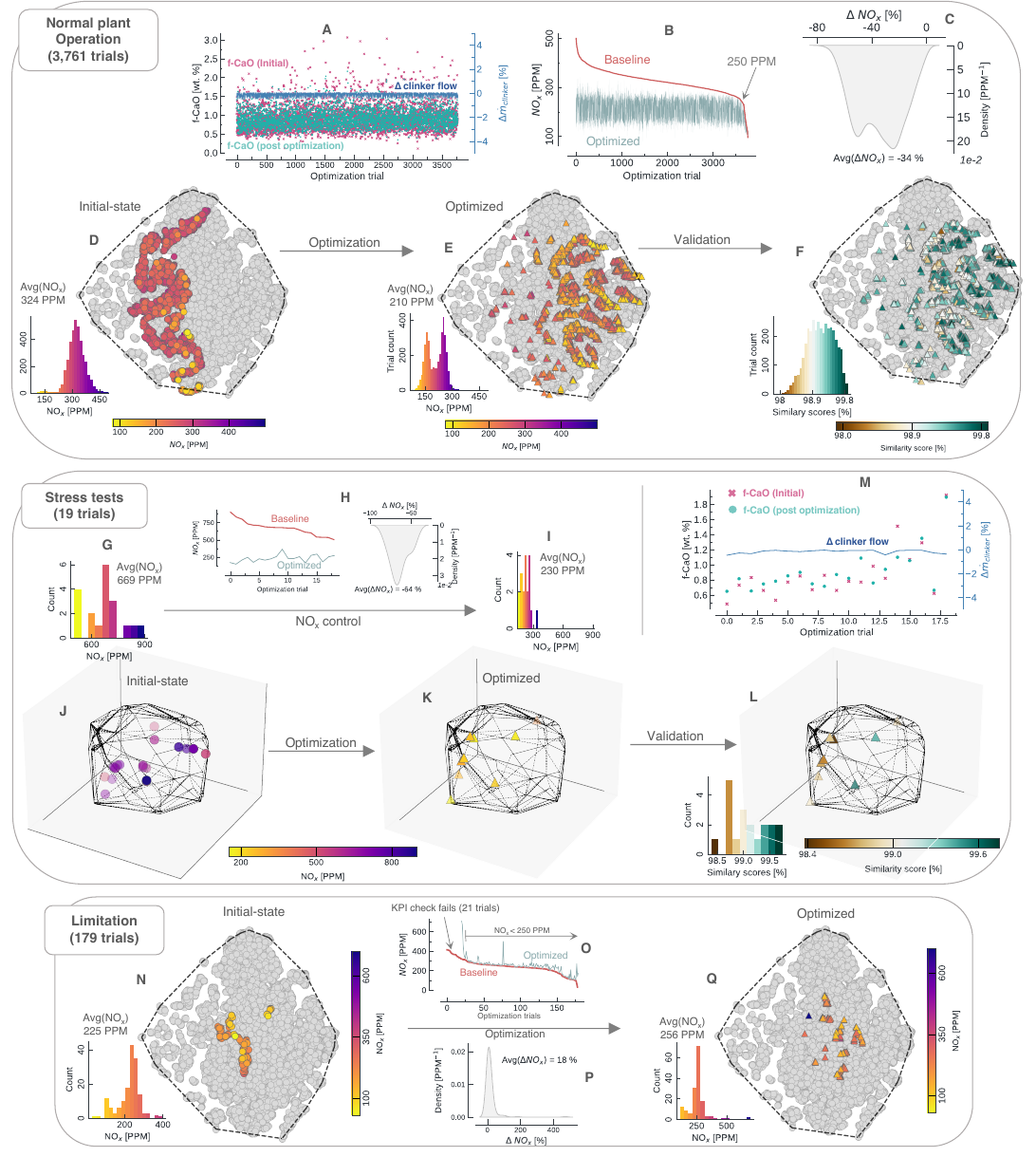}}
\caption{\textbf{Controller performance across different operating scenarios.} A total of 3,761 optimization trials were performed. \textbf{(A-F)} Normal operation trials with initial state $NO\textsubscript{x} \leq 500$ ppm. (A) Comparison of pre vs post--optimization f-CaO (left axis) shows that clinker quality remains unaffected. The right axis shows the relative change in clinker flow ($\Delta$ clinker-flow [\%]) w.r.t the initial state, confirming process stability with deviations close to 0\%. (B) initial state vs optimized NO\textsubscript{x} emissions show that post-optimization values fall mainly within the 100-300 ppm range. For initial-state emissions below $\sim$250 ppm, the controller achieves limited additional reduction, as emissions are already low. (C) On average, the controller achieves a 34\% reduction in NO\textsubscript{x} across all normal-operation trials.
(D-E, J-K, N, Q) visualizes the operational space of the 7 DVs before and after optimization using t-SNE projections. Gray points represent the complete historical plant data, and colored markers correspond to optimization trials. The dotted outline denotes the convex hull of the overall DV space, and marker color encodes the associated NO\textsubscript{x} concentration. (D-E) highlight a shift in the emission distribution from an average of 324 ppm (pre-optimization) to 210 ppm (post-optimization), consistent with an average 34\% reduction (C). (F) shows the Manhattan-distance-based similarity score of the optimized DVs with the nearest data point in the historical plant data. All trials exhibit similarity scores $>$ 98\%, demonstrating that the optimized settings remain within realistic operational boundaries. \textbf{(G-M)} 19 high-stress optimization trials with initial-state NO\textsubscript{x} $>$ 500 ppm). \textbf{(G)} shows the distribution of initial NO\textsubscript{x} (500-1000 ppm) while \textbf{(H-I)} indicates that the optimized emissions converge around an average of 230 ppm, yielding an average reduction of 64\%  with all optimized emissions $<$ 350 ppm. \textbf{(M)} confirms that clinker quality (f-CaO) and flow stability remain unaffected under these conditions.
\textbf{(N-Q)} reports 179 optimization trials to assess the controller limitations. (N) initial-state NO\textsubscript{x} levels (90-430 ppm). (O) controller fails KPI validation for 21 trials, and for the remaining 158 trials where the initial-state $NO\textsubscript{x} <$ 250 ppm, the controller is unable to achieve any reduction in $NO\textsubscript{x}$. (P-Q) Illustrate this effect, with (P) highlighting the net 18\% increase in post-optimization NO\textsubscript{x}. (Q) indicates an average NO\textsubscript{x} rises from 225 ppm to 256 ppm due to the controller.
}
    \label{optimization_fig}
\end{figure}

\noindent To assess the effectiveness of the proposed controller framework under different operational boundaries, we systematically evaluate it across the following scenarios:

\textbf{Scenario 1: Normal plant operation (initial-state NO\textsubscript{x} $\leq$ 500 PPM).}
Under normal operating conditions, the controller was evaluated over 3,761 independent optimization trials. The controller consistently achieved substantial NO\textsubscript{x} mitigation while preserving both clinker quality and production stability (Figs.~\ref{optimization_fig}A-C; Supplementary G6). The optimized operating points exhibit a pronounced shift toward lower emission levels, with most post-optimization NO\textsubscript{x} levels concentrated between 100 and 300 PPM (Fig. \ref{optimization_fig}B). On average, the controller estimates a 34\% reduction in NO\textsubscript{x} concentrations (Fig. \ref{optimization_fig}C). Importantly, for initial-state emissions below ~250 PPM (Fig. \ref{optimization_fig}B), additional optimization yields only marginal improvements, indicating that the plant is already operating near its practical low-emission envelope and offering limited room for further reduction through operational reconfiguration. The high similarity scores (mathematical construction given in Methods) between optimized and historical operating points further confirm that the observed emission reductions are achieved without departing from realistic plant operation (Fig.~\ref{optimization_fig}D-F).

\textbf{Scenario 2: Stress-test (initial-state NO\textsubscript{x} $>$ 500 ppm).}
To check robustness under adverse conditions, the controller was evaluated on 19 stress-test scenarios characterized by elevated initial NO\textsubscript{x} levels exceeding 500 PPM (Fig.~5G). In this regime, the controller demonstrates enhanced effectiveness, driving optimized emissions toward a narrow band centered around 230 PPM (Fig. \ref{optimization_fig}H-I). This corresponds to a 64\% average NO\textsubscript{x} reduction (Fig. \ref{optimization_fig}H), with all optimized emissions remaining below 350 PPM (Fig. \ref{optimization_fig}I). Despite the severity of the initial conditions, clinker quality and throughput stability are preserved (Fig. \ref{optimization_fig}M; Supplementary G6), underscoring the controller's ability to navigate highly non-ideal operating states without violating process constraints. These results highlight the controller's capacity to recover the system from extreme emission excursions and steer operation toward a stable and low-emission regime without departing from realistic plant operation (Fig. ~\ref{optimization_fig}J-L).

\textbf{Controller limitations:} Here, we report 179 optimization trials spanning low-to-moderate initial NO\textsubscript{x} concentrations (Fig.~\ref{optimization_fig}N) for which no net emission reduction is obtained. In a small fraction of cases (21 trials, corresponding to 0.5\% of the 3,959 total control trials performed), KPI validation fails (Fig. ~\ref{optimization_fig}O; Supplementary G6), indicating that no feasible operating point exists that simultaneously reduces NO\textsubscript{x} while preserving clinker quality and production stability. For the remaining 158 trials with initial-state NO\textsubscript{x} $<$ 250 PPM, emission reduction remains constrained by limited optimization headroom, as the plant is already operating close to a local low-emission optimum. In this regime, perturbations of the operating configuration tend to move the system away from this local minimum, resulting in small but measurable increases in NO\textsubscript{x}, with a net average increase of 18\% across this subset (Fig.~\ref{optimization_fig}P-Q). These observations essentially highlight intrinsic process-level limitations rather than flaws of the control methodology.

Altogether, the results demonstrate that the proposed NO\textsubscript{x} controller achieves robust emission reductions across a broad range of operating conditions while consistently maintaining production stability, clinker quality, and operational feasibility. However, the achievable reduction depends on the initial emission level. Greater reductions are observed under high-emission conditions, where there is more room for improvement, whereas improvements are naturally constrained near low-emission operating optima.

\subsection{Techno-economic assessment of the controller}
\label{subsec:economic_benefit}
\noindent To ascertain the economic benefits from the optimization, we perform a high-level estimate by translating the achieved reduction in kiln-inlet NO\textsubscript{x} into avoided SNCR reagent consumption. Since stack-level NO\textsubscript{x} measurements were not available, the analysis assumes proportional translation of kiln-inlet reductions to stack emissions. The actual stack impact may vary depending on plant-specific downstream configurations in the preheater tower (e.g., in-line calciners, staged combustion, or III air injection). Accordingly, plant-specific calibration would be required for precise reagent-savings estimates. Also, the analysis is based on savings from reduced reagent consumption, not on complete SNCR replacement. NO\textsubscript{x} intensity [mg/kg clinker], computed as shown in Supplementary B, is used to calculate NO\textsubscript{x} mass flow [mg/min] = NO\textsubscript{x} intensity [mg/kg clinker] $\times$ clinker flow [kg/min]. Integration NO\textsubscript{x} mass flow [mg/min] and clinker flow [kg/min]over the full two-year dataset, we get:
\begin{equation*}
\text{Annual clinker production of the plant, } 
M_{\mathrm{clk}} = 2.35 \times 10^{6} \text{ t/year},
\end{equation*}
\begin{equation*}
\text{Annual NO\textsubscript{x} emission prior to optimization, } 
M_{NO\textsubscript{x}}^{\mathrm{base}} = 838.39\text{ t/year}.
\end{equation*}
Post-optimization annual emissions are computed using the same minute-level mass-integration procedure. However, at each minute the measured NO\textsubscript{x} concentration is first adjusted according to controller performance, applying a reduction factor $(1-\eta)$, where $\eta=0.34$ when $C_{NO\textsubscript{x}}\leq 500$~PPM and $\eta=0.64$ when $C_{NO\textsubscript{x}}>500$~PPM, before computing the instantaneous NO\textsubscript{x} mass flow. This yields:
\begin{equation*}
\text{Annual NO\textsubscript{x} emission post optimization, } 
M_{NO\textsubscript{x}}^{\mathrm{ctrl}} = 548.16\text{ t/year}.
\end{equation*}
The NO\textsubscript{x} emission intensity normalized to clinker production is defined as $E_{NO\textsubscript{x}} = \frac{M_{NO\textsubscript{x}}}{M_{\mathrm{cl}}}$ $[\mathrm{kg\ NO\textsubscript{x}/t\ clinker}]$. Therefore,
Pre-optimization NO\textsubscript{x} emissions normalized to clinker production,
\begin{equation*}
E_{NO\textsubscript{x}}^{\mathrm{base}} = 0.357\ \mathrm{kg\ NO\textsubscript{x}/t\ clinker}.
\end{equation*}
Post-optimization NO\textsubscript{x} emissions normalized to clinker production,
\begin{equation*}
E_{NO\textsubscript{x}}^{\mathrm{ctrl}} = 0.234\ \mathrm{kg\ NO\textsubscript{x}/t\ clinker}.
\end{equation*}
Reduction in clinker-normalized NO\textsubscript{x} emissions achieved by the controller,
\begin{equation*}
\Delta E_{NO\textsubscript{x}} = 0.123\ \mathrm{kg\ NO\textsubscript{x}/t\ clinker}.
\end{equation*}
Corresponding net annual NO\textsubscript{x} reduction,
\begin{equation*}
\Delta M_{NO\textsubscript{x}} = 290.23\ \mathrm{t/year}.
\end{equation*}
In ammonia-based SNCR systems, NO\textsubscript{x} is reduced primarily through homogeneous gas-phase reactions with injected NH$_3$. The reagent requirement is commonly characterized by the normalized stoichiometric ratio (NSR), defined as the molar ratio of injected NH$_3$ to inlet NO\textsubscript{x} \cite{EPA_SNCR}. For NO\textsubscript{x} reported on an NO$_2$-equivalent mass basis, the ammonia mass required per unit NO\textsubscript{x} removed is $\alpha_{\mathrm{NH_3}} = \frac{\mathrm{NSR}\times M_{NH_3}}{M_{NO_2}}$, where $M_{NH_3}=17.03$~g~mol$^{-1}$ and $M_{NO_2}=46.01$~g~mol$^{-1}$. Practical SNCR operation in cement kilns typically employs NSR values in the range 1.0--1.5 \cite{EUBREF2013,EPA_SNCR}. A conservative value of $\mathrm{NSR}=1.2$ was adopted, yielding $\alpha_{\mathrm{NH_3}} = 0.444\ \mathrm{kg\ NH_3/kg\ NO\textsubscript{x}}$. The ammonia saving per tonne of clinker is therefore
\begin{equation*}
\Delta E_{NH_3}
=
\alpha_{\mathrm{NH_3}} \times \Delta E_{NO\textsubscript{x}}
=
0.0546\ \mathrm{kg\ NH_3/t\ clinker}
\end{equation*}
and the corresponding annual ammonia saving is
\begin{equation*}
\Delta M_{NH_3}
=
\alpha_{\mathrm{NH_3}} \times \Delta M_{NO\textsubscript{x}}
=
128.9\ \mathrm{t/year}
\end{equation*}
Bulk industrial prices of NH\textsubscript{3} is in the ranges of 350--550~USD/t \cite{USGS_AMMONIA, PLATTS_AMMONIA}. Using a mid-range reference price of $p_{NH_3} = 450\ \mathrm{USD/t}$, the resulting monetary savings are $0.0246\ \mathrm{USD/t\ clinker}$, which amounts to $\approx 5.8\times10^{4}\ \mathrm{USD/yr}$.\\

\section*{Conclusion}
The developed framework provides a blueprint for data-driven process optimization that complements ongoing efforts focused on alternative fuels (AF) and novel chemistries. While much of the existing research emphasizes AF or modifications to raw meal (RM), this work highlights a parallel opportunity: optimizing existing production facilities without capital-intensive retrofits or changes to fuels or RM. In this context, we develop an industrial-scale framework for NO\textsubscript{x} control in existing cement plants, leading to the following 4 key outcomes.\\
\indent (i) We curate continuous industrial-scale operational datasets from four cement plants across North America, South America, Europe, and Asia, spanning diverse kiln configurations, fuel mixes, and data richness levels (in both volume and temporal granularity). Leveraging this diversity, we conduct the first global multi-plant benchmark for predicting NO\textsubscript{x}, CO, and CO\textsubscript{2} emissions using 9 machine-learning architectures, where XGBoost consistently achieves the best performance. However, despite identical training protocols, predictive accuracy varies markedly with dataset characteristics: models trained on long-duration, high-frequency datasets (e.g., Plant 1) achieve $\sim$3--5x higher accuracy than those trained on low-volume, coarse-resolution datasets (e.g., Plants 3 and 4). These results indicate that data richness, rather than model complexity alone, remains a key bottleneck in industrial-scale emission modelling. This finding carries important implications because since existing emission models are developed and validated on data from a single facility, implicitly assuming comparable performance across different kiln configurations and fuel mixes. Our results challenge this assumption and highlight the importance of richer datasets and cross-plant validation for robust industrial emission modelling.

\indent (ii) Since kiln operations evolve across multiple time scales, emissions depend not only on instantaneous operating states but also on their temporal evolution. To capture this dynamic behaviour, we incorporate short-term histories of key process variables into the ML models---an aspect largely overlooked in the literature. Accounting for this temporal dependence improves NO\textsubscript{x} prediction accuracy by nearly threefold, underscoring the strong influence of prior operating conditions on NO\textsubscript{x} formation.

\indent (iii) We develop early-warning forecasters that anticipate emission overshoot trajectories by 9 minutes for NO\textsubscript{x}, 6 minutes for CO, and 15 minutes for CO$_2$, providing an actionable buffer for gradual intervention. This addresses a key limitation of conventional kiln control, where corrective actions are typically initiated only after excursions have already occurred.

\indent (iv) Building on these capabilities, we develop a control framework that achieves 34–64\% reduction in NO\textsubscript{x} concentrations across diverse operating conditions while preserving production throughput, clinker quality, and operational feasibility. The controller is validated against the closest historically observed operating states. Applied to Plant~1, producing 2.3~Mt clinker annually, the framework indicates a potential reduction of $\sim$290~t NO\textsubscript{x}/year---equivalent to nearly four months of plant emissions. By suppressing NO\textsubscript{x} formation upstream, the controller also reduces the load on downstream ammonia-based SNCR systems, corresponding to an estimated saving of $\sim$129~t NH$_3$ annually (valued at $\sim$58,000 USD).

Beyond emission mitigation, the framework advances sustainable cement manufacturing by slashing SNCR NH\textsubscript{3} dosings and excess energy consumption. Conventional kiln control often relies on wide safety margins---higher temperatures, air flows, and fuel inputs---to avoid under-burning and ensure complete clinker phase formation, inadvertently increasing energy demand and SNCR NH\textsubscript{3} consumption. In contrast, the proposed controller recommends precise adjustments to controllable decision variables (DVs), enabling optimal air-fuel ratios while avoiding unnecessary temperature elevations and energy waste. Compared with strategies based on modifying fuels or raw feed, which often face adoption inertia, the proposed approach has lower implementation barriers because it relies solely on tuning DVs without altering throughput, product quality, or raw material and fuel supply chains, allowing straightforward deployment in digitally monitored kilns.

Although the framework is not deployed in an operating plant yet, this proof-of-concept establishes the foundation for fully automated digital twins that can enhance existing kilns with minimal structural modifications, potentially accelerating progress toward the UN Net-Zero 2050 sustainable production goals \cite{deutch2020net}.\\

\section*{Methods}
\subsection{Data description and pre-processing}
\label{subsec:data}
\noindent Large-scale operational datasets were curated from four industrial cement plants distributed across North America (Plant~1), South America (Plant~2), Europe (Plant~3), and Asia (Plant~4). The plants differ substantially in operational configuration, location-dependent regulatory compliance with emission limits, fuel mix design, and data features, including volume and temporal resolution, as summarized in Fig.~\ref{plant_compare_fig}A and Table~\ref{plant_compare_table}. Plants~1 and~2 provide approximately two years of data, whereas Plants~3 and~4 provide one year and three months of data, respectively. Sampling frequencies range from 1~min to 2~h, resulting in significant variation in data granularity across datasets. The databases comprise three components:
\begin{itemize}
\item \textbf{Configuration}: Plant specifications including kiln features, bypass systems, and preheater architecture, i.e., number of strings and stages.
\item \textbf{Operation}: Operational details of the plant including calciner and kiln fuel, cyclone-wise temperature profile, pressure profiles, air flow rates, etc.
\item \textbf{Composition}: Material compositions, reported as oxide weight percentages (wt.\%), of the raw meal and clinker.
\end{itemize}
Compositions are quantified using X-ray fluorescence (XRF). Detailed descriptions of the raw datasets for all four plants are provided in Tables~S3--S7 (Supplementary~B). All statistics reported are based on raw plant data. The data exhibits challenges typical of industrial-scale data acquisition, including measurement uncertainty, missing values, duplicated records, and physically inconsistent entries. To ensure data integrity, a uniform four-stage pre-processing protocol\cite{fayaz2025industrial} is used for all plants.
\begin{enumerate}
    \item \textbf{Consistency checks}: Duplicate records are consolidated, and samples with incomplete feature sets are removed to ensure a consistent input space.
    \item \textbf{Physical validation}: Physically implausible values, such as negative compositions and inconsistent XRF--XRD measurements violating thermodynamic constraints, are removed.
    \item \textbf{Outlier removal}: Data is filtered within a range of 0.01-99.99 percentiles. Pre-processing statistics, including data volumes before and after each step, are reported in Table~\ref{plant_compare_table}. Comparisons of raw versus pre-processed distributions for all plants are shown in Figs.~S6--9, demonstrating close agreement between raw and cleaned datasets.
    \item \textbf{Removal of highly correlated features}: To avoid multi-collinearity, features exhibiting intercorrelation above 80\% are pruned. For any group of $n$ highly correlated process parameters, $n-1$ are removed, retaining the variable with the strongest correlation to the target emission. As a result, for example, Plant~4 is reduced from 127 process parameters to 54 relevant inputs. Correlation heatmaps and the systematic feature-reduction procedure for all plants are provided in Figs. S10--13.
\end{enumerate}
This standardized pre-processing framework ensures that observed performance differences across plants arise from intrinsic data characteristics, rather than methodological inconsistencies, thereby enabling a fair and interpretable benchmarking of emission prediction. These datasets were facilitated through the Innovandi consortium.

\subsection*{Evaluation metrics}
\noindent Models are evaluated using the following metrics:\\
(i) Mean Absolute Percentage Error (MAPE)\cite{montano_moreno_using_2013}
\begin{equation}
MAPE=\frac{1}{n}  \sum_{i=1}^{n} \frac{|y_p\ (i)-y_t\ (i)|}{y_t\ (i)}
\end{equation}
where $n$ is the sample size, $y_p(i)$ denotes the predictions and $y_t(i)$ denotes the true values.\\
(ii) Mean absolute error (MAE)
\begin{equation}
MAE=\frac{1}{n}  \sum_{i=1}^{n} |y_p\ (i)-y_t\ (i)|
\end{equation}
(iii) Coefficient of determination $(R^{2})$ \cite{noauthor_pearson_nodate}
\begin{equation}
R^2= 1-\frac{\textnormal{RSS}}{\textnormal{TSS}}\
\end{equation}
where,
\begin{align*}
\textnormal{RSS}= \sum_{i=1}^{n} (y_p(i)-y_t(i))^2 &&
\textnormal{TSS}= \sum_{i=1}^{n} (y_t(i)-\Bar{y}_{t})^2 &&
\bar{y}_{t}=\frac{1}{n} \sum_{i=1}^{n} (y_t(i))
\end{align*}

\subsection{Process-history selection}
\label{subsec:process_history_methods}
\noindent If a EP model uses $n$ process parameters, incorporating a $\tau$-minute process history increases the input dimensionality to $D(\tau)=n(\tau+1)$. Fig. ~\ref{current_vs_process_fig}A illustrates the resulting trade-off between predictive accuracy and model complexity. Increasing $\tau$ reduces MAPE and MAE by capturing delayed emission responses, but rapidly increases feature dimensionality, training time, and GPU memory demand, limiting exploration of longer histories. For all three emissions, error reduction saturates as $\tau$ approaches $\sim$20~min; accordingly, a 20~min process history was adopted for all EPH models.

\subsection{Data pre-processing for forecasting}
\label{subsec:forecast_data}
\noindent The raw emission measurements contain substantial temporal discontinuities due to missing records, sensor dropouts, and scheduled plant shutdowns. These discontinuities are explicitly identified and removed to ensure temporal consistency in the pre-processed dataset, as detailed in Table~S12, Supplementary~F. Two extended gaps, from 13~Jan~2020 - 10~Feb~2020 and from 13~Jan~2021 - 10~Feb~2021, corresponding to annual maintenance shutdowns, are common across all three emissions (Fig.~\ref{forecast_trajectories_fig}A; Supplementary~F1). After pre-processing, the emissions exhibit a stable statistical structure with well-defined means and variances relative to the raw data (Supplementary~F1). The resulting time-series are segmented into continuous temporal segments, which serve as the basis for constructing autoregressive training samples. All pre-processing steps, segment-level statistics, and sample counts are reported in Table~S12, Supplementary~F1.

\subsection{Look-back window selection}
\label{subsec:lookback_selection}
\noindent Single-step forecasters are evaluated over look-back windows ranging from 5 minutes to 2 hours, at 5-minute intervals. Forecast accuracy, quantified using MAPE and MAE, improves with increasing window length but exhibits diminishing returns beyond a threshold, reflecting a trade-off between temporal coverage and model complexity (Fig.~\ref{forecast_trajectories_fig}D). Based on this analysis, a 25-minute look-back window is selected as the optimal balance between accuracy and computational efficiency and is used for all forecasting models.

\subsection{Forecast methodology}
\label{subsec:forecast_method}
\noindent  Emission forecasting is formulated as a univariate auto-regressive (AR) problem, in which future emissions are predicted solely from their past observations. Let $\{y_t\}_{t=1}^{T}$ denote the emission time-series sampled at one-minute resolution. Given a look-back window of length $L$, the AR input vector at time $t$ is defined as $\mathbf{x}_t = \left[y_{t-L}, y_{t-L+1}, \ldots, y_{t-1}\right]^\top $. Two forecasting approaches are compared in this work:

\textit{(i) Recursive single-step forecaster}: predicts the emission value one minute ahead using the selected 25-minute look-back window, learning a mapping
\begin{equation}
\hat{y}_{t+1} = f_{\text{SS}}(\mathbf{x}_t),
\end{equation}
where $f_{\text{SS}}(\cdot)$ predicts the emission one minute ahead; SS denotes 'single step' formulation. The single-step forecaster is trained by minimizing the forecast error
\begin{equation}
\mathcal{L}_{\text{SS}} =
\frac{1}{N} \sum_{t}
\ell \!\left( y_{t+1}, \hat{y}_{t+1} \right),
\end{equation}
where $\ell(\cdot)$ denotes the pointwise loss function (mean squared error), and $N$ denotes the number of training samples. Forecasts over a horizon $H$ are obtained recursively:
\begin{equation}
\hat{y}_{t+k} =
f_{\text{SS}}\!\left([\hat{y}_{t+k-L}, \ldots, \hat{y}_{t+k-1}]\right),
\quad k = 1,\ldots,H,
\end{equation}

\textit{(ii) Multi-step forecaster}: is trained to directly predict the entire emission trajectory over a predefined horizon ($H$) in a single forward pass:
\begin{equation}
\left[\hat{y}_{t+1}, \hat{y}_{t+2}, \ldots, \hat{y}_{t+H}\right]^\top
= f_{\text{MS}}(\mathbf{x}_t).
\end{equation}
Training minimizes a horizon-aggregated loss function, which explicitly penalizes prediction errors across the entire forecast horizon.
\begin{equation}
\mathcal{L}_{\text{MS}} =
\frac{1}{N} \sum_{t}
\sum_{k=1}^{H}
\ell \!\left( y_{t+k}, \hat{y}_{t+k} \right),
\end{equation}
\subsection{Error propagation}
\noindent For recursive single-step forecasters, prediction errors accumulate through repeated model application. Denoting the one-step prediction error as $\varepsilon_{t+1} = y_{t+1} - \hat{y}_{t+1}$, the $k$-step-ahead error can be expressed recursively as
\begin{equation}
\varepsilon_{t+k}
\approx
\sum_{i=1}^{k}
\left( \prod_{j=i+1}^{k} J_{t+j} \right)
\varepsilon_{t+i},
\end{equation}
where $J_{t+j}$ represents the local sensitivity (Jacobian) of the forecaster with respect to its inputs. This expression highlights error amplification as iterations increase. In contrast, a multi-step forecaster estimates all future values jointly, avoiding recursive feedback, thereby mitigating systematic error growth over the prediction horizon. Both forecasting formulations are trained and tuned using identical data splits and are evaluated over a common 60-minute forecast horizon ($H=60$). Training details for the single-step and multi-step forecasters are provided in Supplementary F2. Comparative performance on the train vs. test sets is reported in Supplementary F3-F4.

\subsection*{Penalties for promoting operational plausibility}
\label{sec:realism_penalties}
\noindent  To prevent the optimizer from proposing control configurations that are statistically inconsistent with feasible kiln operation, two Mahalanobis distance-based penalties are introduced in the loss function of the optimizer:

\textbf{1. Correlation consistency penalty} preserves historically observed multivariate process correlations. It is formulated as a Mahalanobis distance in standardized decision-variable (DV) space:
\begin{equation}
P_{\mathrm{corr}}(\mathbf{x})
=
\bigl(\mathbf{z}_{\mathrm{corr}}-\boldsymbol{\mu}_{\mathrm{corr}}\bigr)^{\top}
\mathbf{C}_{\mathrm{corr}}^{-1}
\bigl(\mathbf{z}_{\mathrm{corr}}-\boldsymbol{\mu}_{\mathrm{corr}}\bigr),
\label{eq:P_corr}
\end{equation}
where $x \in \mathbb{R}^{p}$ is the candidate DV vector, $z_{\text{corr}}$ is the standardized version of $x$, $\mu_{\text{corr}}$ is the empirical mean of the historical DV dataset, $C_{\text{corr}} = \begin{bmatrix}
\sigma_1^2 & \rho_{12}\sigma_1\sigma_2 & \cdots \\
\rho_{21}\sigma_2\sigma_1 & \sigma_2^2 & \cdots \\
\vdots & \vdots & \ddots
\end{bmatrix}$ is the covariance matrix, where $\sigma_i^2$ denotes the variance of DV $i$, and $\rho_{ij}$ denotes the Pearson correlation coefficient between DVs $i$ and $j$. The diagonal elements quantify the dispersion of individual DVs, while the off-diagonal elements encode historically observed
co-variation between DVs. Since the deviation is scaled by $C_{\text{corr}}^{-1}$, the penalty measures distance relative to the historical covariance structure rather than absolute magnitude. Candidate solutions violating the historically observed co-variation pattern incur a large Mahalanobis penalty. However, plants operate under multiple process conditions, resulting in regime-dependent correlation patterns among DVs. Since C\textsubscript{corr} is estimated from the entire historical data spanning all such regimes, it captures the combined variability and co-variation pattern exhibited by DVs across these operating states. Therefore, the correlation penalty does not impose a single fixed relationship between variables. Instead, it constrains the optimization to remain statistically consistent with the historically observed regime-dependent operating patterns.

\textbf{2. Operational feasibility penalty} constrains optimized set points to remain close to the historical operating manifold. Let $\mathbf{z}_{\mathrm{operate}}\in\mathbb{R}^{n}$ denote the standardized control vector using a scaler fitted to the full operating dataset, with empirical mean $\boldsymbol{\mu}_{\mathrm{operate}}$ and covariance $\mathbf{C}_{\mathrm{operate}}$. The penalty is defined as
\begin{equation}
P_{\mathrm{operate}}(\mathbf{x})
=
\bigl(\mathbf{z}_{\mathrm{operate}}-\boldsymbol{\mu}_{\mathrm{operate}}\bigr)^{\top}
\mathbf{C}_{\mathrm{operate}}^{-1}
\bigl(\mathbf{z}_{\mathrm{operate}}-\boldsymbol{\mu}_{\mathrm{operate}}\bigr),
\label{eq:P_real}
\end{equation}
thereby discouraging control combinations that lie outside the envelope of historically feasible kiln operation. Both penalties are incorporated into the objective function with scalar weights $W_{\mathrm{corr}} = 0.20$ and $W_{\mathrm{operate}} = 0.15$.

\subsection*{Optimization constraints}
\label{constraints_section}

\textbf{1. Raw meal (RM) constraint.} RM flow rate and chemistry are governed by quarry planning, material availability, blending logistics, and clinker quality control. Modifying these parameters is neither trivial nor rapidly implementable in practice. Changes in RM chemistry require availability of suitable corrective materials, which may not be practically feasible for a given plant at a given time. Additionally, RM adjustments propagate slowly through the preheater-kiln system due to long material residence times. We also intentionally avoid modifying RM flow rate to ensure that clinker production targets remain unaffected. For these reasons, RM flow rate and chemistry are held fixed during optimization. Instead, only operational decision variables (air flows, fuel rates, draft conditions), which can be directly and reliably adjusted by plant operators, are considered as control levers for NO\textsubscript{x} mitigation.

\textbf{2. Decision variable constraint.} The initial population is generated by sampling each decision variable within a bounded interval of $\pm5\%$ around its initial value $x_i^{0}$:
\begin{equation}
x_i^{\mathrm{lower}} = x_i^{0}(1-\delta),
\qquad
x_i^{\mathrm{upper}} = x_i^{0}(1+\delta),
\qquad
\delta = 0.05,
\label{eq:bounds}
\end{equation}
\begin{equation}
x_i^{(0)} \sim \mathcal{U}\!\left(x_i^{\mathrm{lower}},\,x_i^{\mathrm{upper}}\right),
\qquad i=1,\ldots,n,
\label{eq:initial_sampling_1}
\end{equation}
where $\mathcal{U}$ denotes the continuous uniform distribution.\\

\textbf{3. Total fuel constraint.} Total fuel input is constrained not to exceed the initial operating level. Also, note that the fuel composition is constant in this study. Denoting kiln and calciner fuel rates as $x_{\mathrm{kiln}}$ and $x_{\mathrm{calciner}}$, respectively, the constraint is
\begin{equation}
x_{\mathrm{kiln}} + x_{\mathrm{calciner}}
\le
x_{\mathrm{kiln}}^{0} + x_{\mathrm{calciner}}^{0}.
\label{eq:total_fuel_constraint}
\end{equation}
In the numerical implementation, violation of any hard constraint results in assigning a penalty of order $10^{6}$ to the objective function, effectively eliminating such candidates during selection.

\subsection*{Manhattan distance based similarity scores for validating the NO\textsubscript{x} controller}
\noindent Each optimized solution is validated against historical plant operations using a Manhattan-distance-based similarity analysis. Specifically, let $\mathbf{x}^{\ast}=(x^{\ast}_1,\dots,x^{\ast}_p)$ denote an optimized operating point and $\mathbf{x}^{(k)}=(x^{(k)}_1,\dots,x^{(k)}_p)$ the $k$th historical point, where $p$ is the number of decision variables. To ensure comparability across variables with different units, all variables are normalized using min-max scaling derived from the historical dataset as $\tilde{x}_j=\frac{x_j-x_j^{\min}}{x_j^{\max}-x_j^{\min}}, \quad j=1,\dots,p.$ The Manhattan distance between the optimized point and each historical point in normalized space is computed as:
\begin{equation}
D_{\mathrm{Man}}^{(k)}
=
\sum_{j=1}^{p}
\left|
\tilde{x}^{(k)}_j - \tilde{x}^{\ast}_j
\right|,
\label{eq:manhattan_distance}
\end{equation}
and the nearest historical operating point is identified as $k^{\ast}=\arg\min_k D_{\mathrm{Man}}^{(k)}$. This distance is converted into an interpretable similarity score, i.e.,
\begin{equation}
S
=
\left(
1 - \frac{D_{\mathrm{Man}}^{(k^{\ast})}}{p}
\right)
\times 100\%,
\label{eq:similarity_score}
\end{equation}
representing the average per variable agreement between the optimized solution and the most similar historical state. High similarity scores indicate that the optimized decision variable configurations lie within the plant's empirical operating envelope, confirming the practical deployability of the proposed NO\textsubscript{x} controller.

\subsection*{Shapley additive explanations (SHAP)}
\hypertarget{shap section}{}
\noindent To interpret the input-output relationships learned by complex ML models, we employ SHapley Additive exPlanations (SHAP) \cite{lundberg_unified_2017}. The SHAP attribution $\phi_k$ associated with feature $k$ quantifies its average marginal contribution to the model prediction over all possible feature coalitions \cite{bansal2025interpretable}. Formally, the attribution is defined as
\begin{equation}
\phi_k
=
\sum_{\mathcal{S} \subseteq \mathcal{F} \setminus \{k\}}
\frac{|\mathcal{S}|!\,(d-|\mathcal{S}|-1)!}{d!}
\left[
g(\mathcal{S}\cup\{k\}) - g(\mathcal{S})
\right],
\label{eq:shap_definition}
\end{equation}
where $g(\cdot)$ denotes the trained predictive model, $\mathcal{F}$ is the full set of $d$ input features, and $\mathcal{S}$ represents a subset of features excluding $k$. $g(\mathcal{S})$ denotes the model output when only features in $\mathcal{S}$ are present, while $g(\mathcal{S}\cup\{k\})$ denotes the output after including feature $k$ \cite{waris2024pseudo}. This formulation ensures a fair allocation of feature contributions by averaging marginal effects across all possible feature orderings.\\

\section*{Data availability}
The datasets analyzed in this work were provided by industrial collaborators through the Global Cement and Concrete Research Network (GCCRN) under a formal confidentiality agreement. Due to proprietary and contractual obligations, the underlying raw data cannot be made publicly accessible. Requests for additional information may be addressed to the corresponding author and will be considered subject to consent from GCCRN.\\

\section*{Code availability}
The source code for developing the ML models for this study is publicly available at the following repository:[\url{https://doi.org/10.5281/zenodo.18888654}]\cite{sheikhjunaidfayaz_2025_15302728}\\

\section*{Computational environment}
All computations were conducted on a Linux machine running Ubuntu 22.04 with Python 3.9.12. The system was equipped with an Intel\textsuperscript{\textregistered} Xeon\textsuperscript{\textregistered} Gold 6226R CPU (32 physical cores, 64 threads) and 62.5 GB of RAM. GPU-accelerated models, such as XGBoost, were trained using an NVIDIA RTX A2000 12 GB GPU with driver version 580.65.06 and CUDA version 13.0, utilizing the gpu\_hist tree method.\\

\bibliographystyle{elsarticle-num}
\bibliography{References_AUTOCEM}

\section*{Acknowledgements}
The authors acknowledge the computational resources provided by HPC IIT Delhi, as well as the financial support and assistance with data acquisition from Innovandi--The Global Cement and Concrete Research Network.\\

\section*{Author contributions}
N.M.A.K. and S.J.F. conceptualized of the study. N.M.A.K. and S.J.F. developed the methodology including the machine learning framework. N.D.M.B. and S.J.F. performed data preprocessing. S.J.F. implemented codes, trained the models, generated the visualizations, and prepared the original draft of the manuscript. N.M.A.K., M.G., M.R., and S.B. were responsible for funding acquisition, project administration, resource provision, and overall supervision of the work. All authors contributed to the analysis of the results, interpretation and reviewing and editing the manuscript.
\\

\section*{Competing interests}
The authors declare no competing interests.\\

\section*{Ethical approval}
This study was carried out with approval from the Global Cement and Concrete Research Network (GCCRN), which provided financial support and facilitated data access. All datasets were anonymized to prevent the disclosure of confidential or plant-specific information related to the participating cement facilities.\\

\subsection*{Correspondence} 
\noindent All queries on the study should be addressed to N. M. Anoop Krishnan.

\setcounter{section}{0}
\renewcommand{\thesection}{\arabic{section}}
\newpage
\end{document}